\documentclass{ifacconf}

\usepackage{graphicx}
\usepackage{natbib}
\usepackage{amsmath}
\usepackage{amssymb}
\usepackage{caption}
\usepackage{subcaption}

\usepackage{color}
\begin{document}
\begin{frontmatter}

\title{Probabilistic Height Grid Terrain Mapping for Mining Shovels using LiDAR} 

\author{Vedant Bhandari\hspace{2mm}} 
\author{Jasmin James\hspace{2mm}} 
\author{Tyson Phillips}
\author{P. Ross McAree}

\address{The School of Mechanical and Mining Engineering, The University of Queensland, 4072, Australia (e-mail: v.bhandari@uq.edu.au).}

\tiny{© 2024 the authors. This work has been accepted to IFAC for publication under a Creative Commons Licence CC-BY-NC-ND”.
When the article is published, the posted version should be updated with a full citation to the original IFAC-PapersOnline publication, including DOI.}

\begin{abstract}
    This paper explores the question of creating and maintaining terrain maps in environments where the terrain changes.
    The specific example explored is the construction of terrain maps from 3D LiDAR measurements on an electric rope shovel.
    The approach extends the height grid representation of terrain to include a Hidden Markov Model in each cell, enabling confidence-based mapping of constantly changing terrain.
    There are inherent difficulties in this problem, including semantic labelling of the LiDAR measurements and determining the pose of the sensor.
    The significance of this work lies in the need for accurate terrain mapping to support autonomous machine operation.
\end{abstract}

\begin{keyword}
    Robotics, automation, modelling, perception and sensing, map building.
\end{keyword}

\end{frontmatter}
\section{Introduction}
Rope shovels are used to extract and load overburden and ore from blasted terrain.
Automating shovel operation is expected to provide numerous economic and environmental benefits.
Making autonomous decisions requires a perception system to construct and develop an accurate representation of the workspace.
The perception tasks typically include end-effector tracking for collision avoidance, identifying other agents in the workspace, such as haul trucks and clean-up equipment, and generating a map of the environment.
The quality of the inputs influences the optimality of the subsequent decision-making processes.

In this paper, we focus on generating a terrain map using a shovel-mounted 3D Light Detection and Ranging (LiDAR) sensor.
A terrain map is required to support real-time excavation sequencing, manage the interaction between the dipper and the dig face (knowing where and how to initiate the dig), and allow productivity monitoring.
The contributions of this work are to (i) provide an overview of a terrain mapping framework using a shovel-mounted LiDAR, (ii) detail a novel probabilistic approach to update a terrain map using LiDAR measurements, and (iii) demonstrate the approach using field datasets.   

Generating a terrain map from a shovel-mounted sensor is a challenging task.
A 3D LiDAR sensor provides a set of range measurements at specified heading and elevation angles.
The sensor has range measurement uncertainty and requires intrinsic and extrinsic calibration to provide accurate point cloud data, see~\cite{Nou:16,Dad:18}.
The sensor's pose is also required to locate point cloud measurements within a global frame.
Errors in the sensor pose propagate to form incorrect beliefs about the environment, see~\cite{Yin:24}.

\begin{figure}[t!]
    \begin{center}
    \includegraphics[width=8.4cm]{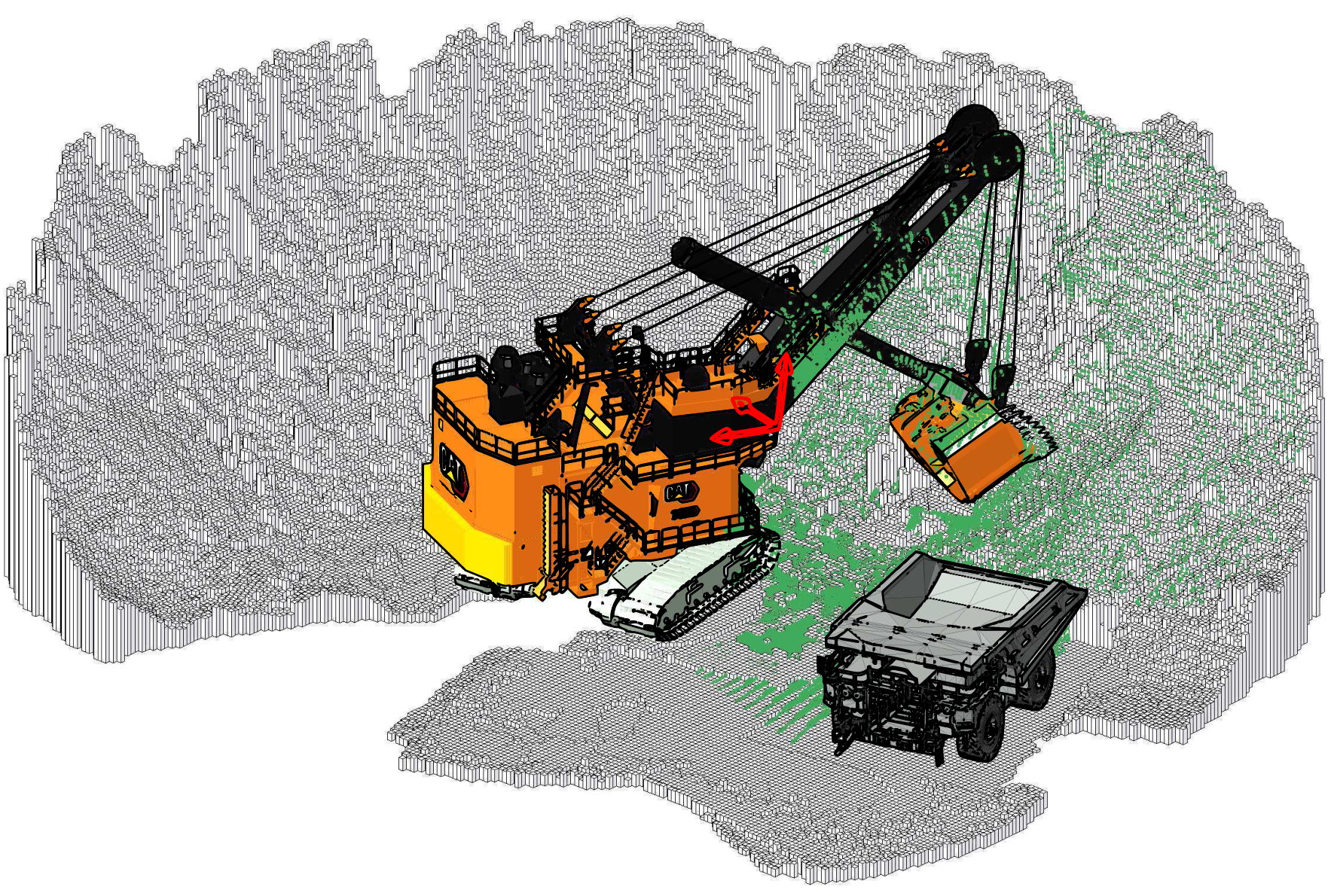}
    \caption{A terrain map generated using the proposed approach with point clouds from a 3D LiDAR mounted to the shovel's house represented by the red coordinate frame.
    An example LiDAR scan is shown in green, with models of the shovel and truck.} 
    \label{fig:shovel_scene}
    \end{center}
\end{figure}

Figure.~\ref{fig:shovel_scene} illustrates the shovel's operating environment and the data captured from a shovel-mounted LiDAR.
These measurements provide no semantic information and require interpretation to generate beliefs about the environment they represent.
This is challenging as the point cloud captures measurements on the shovel, other machines, the terrain, and dust.
The terrain is also dynamically changing, with material being removed and spilled in the workspace.
Dusty environments and material spillage from the dig face and dipper lead to spurious measurements and an unpredictable representation of the environment, see~\cite{Phi:2017}.
A robust approach is needed to map terrain in these challenging conditions.

The paper is organized as follows.
Section~\ref{sec:related_work} provides an overview of related work in extracting static scenes from a dynamic environment and recent terrain mapping methods.
The proposed approach is detailed in Section~\ref{sec:methdology}, with experimental results provided in Section~\ref{sec:results}.
Concluding remarks and future work is identified in Section~\ref{sec:conclusions_and_future_work}.

\section{Related work}
\label{sec:related_work}
This section explores existing work on representing dynamic environments and previous studies on mapping terrain using excavator-mounted sensors.
Gaps are highlighted in the mapping process to support excavation sequencing and real-time productivity monitoring.

Literature is rich with approaches to map static environments using point cloud data.
Occupancy-based methods discretize the environment into voxels (or pixels in 2D) and label each voxel as occupied if a measurement is coincident with the voxel, or free if the voxel intersects the beam's path.
However, these approaches are based on the assumption of a static environment and typically provide accurate results in indoor experiments only, see~\cite{Thr:98,Thr:00}.

Octomap by~\cite{Hor:2013} is a popular framework that updates voxels probabilistically.
First, a new scan is discretized, and each voxel is assigned a probability of being free and occupied.
These probabilities are fused into a global map using the update laws introduced by~\cite{Mor:85}. 
Upper and lower bounds on the occupancy probabilities of each voxel handle adaptation to dynamic objects.
However, changing state requires an equal number of observations to that used to form the belief initially.
This is problematic for the terrain mapping problem as the terrain is mapped static over an extended period but requires quick adaptation to integrate excavated material into the global map.
Configuring lower and upper bounds that support fast transition is not robust because of noise in the measurements.
Voxel update approaches such as exponential forgetting methods explored by~\cite{Ygu:08} experience similar problems.
Variants of Octomap handle dynamic objects by augmenting deep-learning object detection networks to the original framework, see~\cite{Zha:18, Liu:19}.

A height grid representation is used by~\cite{Sou:13} as an alternative volumetric representation to voxels.
A 2D grid composed of cells represents the environment.
Each cell has a height described by a mean and standard deviation, updated using a probabilistic law derived from the Kalman filter formulation.
As before, there is no handling of dynamic changes in the environment.
Recent approaches to mapping a static environment, such as Removert~\citep{Kim:20}, ERASOR~\citep{Lim:21}, and DynamicFilter~\citep{Fan:22}, aim to extract a static map of the environment using free space constraints to update the global map.
The terrain mapping problem is different as the \emph{static} environment changes with the extraction and spillage of material.

These ideas have been used to demonstrate terrain mapping from shovel-mounted sensors.
\cite{Dun:06} generate a Digital Terrain Map (DTM) using a 2D laser scanner mounted to a 1/6 scale rope shovel to image the terrain when the shovel is swinging or in tuck.
Using the DTM for autonomous excavation is listed as future work, and there are no details about handling scenarios with dust and other agents in the workspace.
\cite{Hil:15} generates DTMs by fusing Radio Detection and Ranging (RaDAR) and LiDAR sensing technologies.
The Global Navigation Satellite Systems (GNSS) pose of the shovel is used to stitch successive point clouds, and haul trucks are labelled by segmenting points estimated to be associated with the truck, located by its GNSS pose.
The authors note the sensitivity of the approach to the uncertainty in the shovel and truck pose estimates.
\cite{Bet:23} demonstrates a terrain mapping method using two excavator-mounted 3D LiDARs.
The approach constructs a height grid representation of the dig face using a visibility-based approach where measurements in previously identified free space are disregarded.
While this is generally not a problem in excavation scenarios, material added to the workspace (e.g., face collapses or the movement of material) cannot be modelled.

\section{Methodology}
\label{sec:methdology}
\begin{figure*}[ht!]
    \begin{center}
        \includegraphics[width=\linewidth]{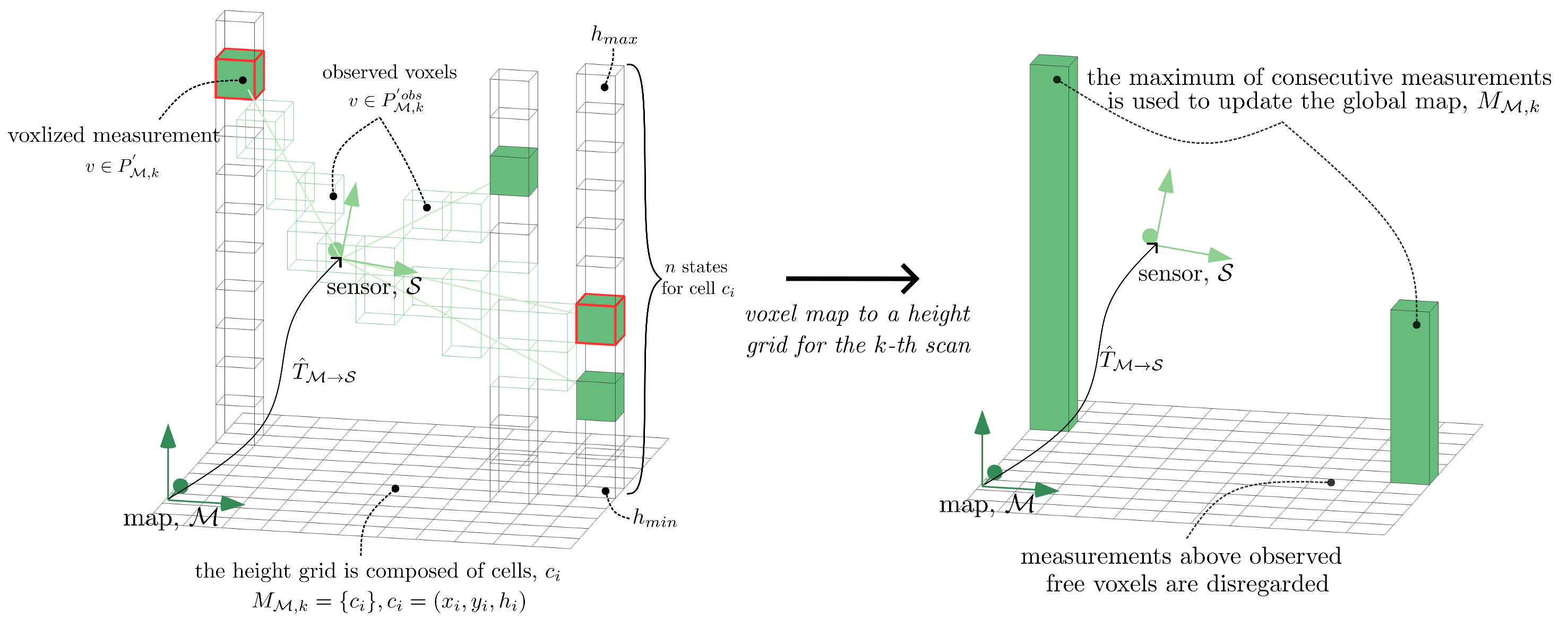}
        \caption{The conversion of the voxelized map (left) to a height grid (right) constructed using the voxelized measurements, \smash{$P_{\mathcal{M},k}^{'}$}, and the raycast result to find all the observed voxels, \smash{$P_{\mathcal{M},k}^{'obs}$}. For each observed cell, \smash{$c_{i}$}, the maximum of consecutive measurements is used to update the global map, \smash{$M_{\mathcal{M},k}$}. Measurements in \smash{$P_{\mathcal{M},k}^{'}$} that are above observed free voxels in \smash{$P_{\mathcal{M},k}^{'obs}$} are disregarded as they are likely to be noisy measurements from dust or missed in the semantic labelling process.}
    \label{fig:methodology_convert_voxel_map_to_height_grid}
    \end{center}
\end{figure*}
The following section details the proposed approach to represent and update a terrain map using a shovel-mounted 3D LiDAR.
The novelty of the method is the ability to update dynamic changes quickly and efficiently to provide a real-time terrain map.
The mathematical foundation of the general HMM framework applied to this problem is adapted from~\cite{Jam:2024}.

\subsection{Map Representation}
A global map frame, $\mathcal{M}$, is established to indicate the origin of the environment.
The map at time $k$, $M_{\mathcal{M},k}$, is represented using a height grid composed of cells, $c$, discretized at a user-defined size of $\Delta$.
The $i$-th cell, $c_{i}$, is represented using a ($x_{i},y_{i}$) coordinate relative to the origin, and a corresponding height, $h_{i}$.
A height grid representation is employed due to its ability to efficiently represent large areas as previously demonstrated by~\cite{Bet:23, DAd:23}.

The aim is to interpret new scans and update the cells in the global map to represent the changing terrain.
A naive approach is to update cells based on new observations and directly overwrite existing measurements.
Changes in the state of the terrain are determined by computing the difference in cell heights between two instances in time.
However, this does not account for any uncertainties in the measurements, sensor pose, and poor sensing conditions due to atmospheric obscurants such as dust.
Hence, using this approach results in errors in the representation of the terrain which can become significant when using differencing to identify changes in terrain.

We propose using probabilistic approaches to integrate the observations into the global map to account for various sources of uncertainty.
Each cell is represented using a Hidden Markov Model (HMM) as the true height of each cell cannot be measured directly.
Using a HMM~\citep{Rab:89} requires defining: the $n$ states of each cell, $S=\{S_{1},\ldots,S_{n}\}$, the transition probabilities between the states encoded in the state transition matrix, $\mathbf{A}\in\mathbb{R}^{n\times n}$, and a process to generate the likelihood of being in each state for the $i$-th cell at time $k$, captured in the diagonal matrix $\mathbf{B}_{i,k}\in\mathbb{R}^{n\times n}$.

Let the $i$-th cell's probability of being in each state at time $k$ be represented by the state vector, $\hat{\mathbf{x}}_{i,k}\in\mathbb{R}^{n\times 1}$.
The cell's state is recursively updated using the HMM filter~\citep{Ell:08},
\begin{equation}
    \label{eq:hmm_filter}
    \hat{\mathbf{x}}_{i,k} = \eta_{i,k}\mathbf{B}_{i,k}\mathbf{A}\hat{\mathbf{x}}_{i,k-1},
\end{equation}
where $\eta$ is a normalization constant to ensure the state vector, $\hat{\mathbf{x}}_{i,k}$, is a probability.
The initial state is given by $\hat{\mathbf{x}}_{i,0}$.

The possible height range of each cell, $[h_{min},h_{max}]$, is discretized to intervals of size $\Delta$, to form the $n$ states of the HMM,
\begin{equation}
    n = \bigg\lfloor\frac{h_{max}-h_{min}}{\Delta}\bigg\rfloor + 1,\hspace{1mm} n\in\mathbb{Z}.
\end{equation}
The recursive state update process described in (\ref{eq:hmm_filter}) allows for the efficient update of large state vectors.
The ($n\times n$) state transition matrix is set as,
\begin{equation}
    \mathbf{A} =\begin{bmatrix}
                    a_{11} & a_{12} & \dots & a_{1n} \\
                    a_{21} & a_{22} & \dots & a_{2n} \\
                    \vdots & \vdots & \ddots &  \vdots \\
                    a_{n1} & a_{n2} & \dots & a_{nn}
                 \end{bmatrix} =  
                \begin{bmatrix}
                    0.99 & \delta & \dots & \delta \\
                    \delta & 0.99 & \dots & \delta \\
                    \vdots & \vdots & \ddots & \vdots \\
                    \delta & \delta & \dots & 0.99
                 \end{bmatrix},
\end{equation}
where,
\begin{equation}
    \delta = \frac{1-0.99}{n-1},
\end{equation}
to ensure each row and column sum to one.
The self-transition probability of each state is set to a large value of $0.99$ to allow for the accumulation of sufficient evidence before changing state.
The off-diagonal terms are set to low values of $\delta$ to allow for transitions from any state.
Each observed cell is initialized in the state corresponding to its first measurement.
Erroneous initialization is handled by allowing $m$ observations to incrementally update the global map before using it for subsequent planning stages.
The global map may also be initialized by using a survey of the area to be mapped.

\subsection{Map Update}
This section describes the process of integrating a new LiDAR observation at time $k$ into the map, $M_{\mathcal{M},k}$.

The $k$-th point cloud in the sensor frame ($\mathcal{S}$), $P_{\mathcal{S},k}$, is semantically labelled to estimate point cloud measurements corresponding to the shovel, the truck being loaded, and any other agents in the environment.
An assumption is made that the remaining measurements correspond to the terrain. 
Due to the controlled operating environment, we find this simplifying assumption to hold.

The shovel is labelled by fitting known geometries of the house, boom, dipper assembly (handle, dipper, dipper door), and tracks at their estimated poses relative to the sensor and labelling all measurements close to these geometries.
The boom and house are static relative to the sensor and are labelled using geometric constraints.
The pose of the dipper assembly and tracks are estimated using pose estimation algorithms that provide sufficiently accurate performance, e.g. MSoE~\citep{Phi:18} or PLuM~\citep{Bha:23}.
Methods to validate this semantic labelling, e.g.~\cite{Phi:16}, are also useful in this process.
Machine data can be alternatively used if available.
Other agents in the workspace, such as the trucks and clean-up equipment, can be identified by locating their GNSS pose relative to the sensor or using the geometry pose estimation algorithms mentioned previously.
Spurious measurements from dust are also labelled as terrain and handled in the process described below.

The semantically labelled scan is located in the map frame, $P_{\mathcal{M},k}$, using the sensor's pose estimate, $\hat{T}_{\mathcal{M}\rightarrow\mathcal{S},k}$,
\begin{equation}
    P_{\mathcal{M},k} = \hat{T}_{\mathcal{M}\rightarrow\mathcal{S},k} P_{\mathcal{S},k}.
\end{equation}
The sensor's pose is estimated using a Kalman filter to fuse measurements from GNSS and LiDAR odometry approaches such as SiMpLE~\citep{Bha:24}.
The point cloud in the map frame, $P_{\mathcal{M},k}$, is discretized at a resolution of $\Delta$ to form voxelized scan $P_{\mathcal{M},k}^{'}$.
The measurement $z$-values from the scan in the map frame, $P_{\mathcal{M},k}$, are preserved for each voxel, $v$, to avoid quantization errors in the following steps.

A process is needed to convert the voxelized map to a height grid where each ($x,y$) coordinate is assigned a height.
The process used assumes that terrain must extrude from the ground and each cell must only have a single height.
This assumption does not hold for terrain with overhangs and is a known limitation of the selected height grid representation~\citep{Bet:23}.
Figure.~\ref{fig:methodology_convert_voxel_map_to_height_grid} illustrates the process to aid the explanation below.
First, a raycast is performed from the sensor's pose to each voxelized measurement in $P_{\mathcal{M},k}^{'}$ to find all the observed voxels, $P_{\mathcal{M},k}^{'obs}$.
All voxels in the LiDAR's beam path from the sensor to the measurement are labelled as free and the measurement is labelled as occupied.
For the $i$-th cell, $c_{i} \in P_{\mathcal{M},k}^{'}$, all voxelized observations for the $z$-coordinate in $P_{\mathcal{M},k}^{'obs}$ are saved in order of increasing height, $H_{i}=\{h_{0}, h_{1}, \ldots, h_{j}\}$.
If $h_{0}$ is a free voxel, the cell is removed from the list of cells to be updated as terrain cannot exist above the observed space as assumed previously.
This constraint avoids updating measurements corresponding to dust or measurements missed from the semantic labelling process.
If $h_{0}$ is labelled as occupied, the height is incremented until a free space voxel or the last element of $H$ is reached.
The unquantized height corresponding to this value, $h_{i}$, is assigned to cell $c_{i}$.
This process is repeated for all cells to construct the height grid for this scan, $P_{\mathcal{M},k}^{h}= \{c_{i}\}$.

The height, $h_{i}$, for each cell, $c_{i}$, is used to generate the likelihood of being in each of the $n$ states at time $k$, and construct the ($n\times n$) diagonal likelihood matrix,
\begin{equation}
    \small
    \mathbf{B}_{i,k} = \begin{bmatrix}
                        \mathcal{N}(h_{i}|h_{S_{1}},\sigma) & 0 & \dots & 0 \\
                        0 & \mathcal{N}(h_{i}|h_{S_{2}},\sigma) & \dots & 0 \\
                        \vdots & \vdots & \ddots & \vdots \\
                        0 & 0 & \dots & \mathcal{N}(h_{i}|h_{S_{n},\sigma})
                       \end{bmatrix},
\end{equation}
where $\sigma$ is a user-configured standard deviation used to represent the uncertainty in the cell's height estimate.
The state of the global map, $M_{\mathcal{M},k}$, is then updated using (\ref{eq:hmm_filter}).
The $i$-th cell's height (state) is updated when the probability of being in that state exceeds a predefined threshold, $p_{min}$, else the cell remains in its current state,
\begin{equation}
    S_{i,k} = S\Big(\underset{i}{\arg\!\max}(\hat{\mathbf{x}}_{i,k})\Big) | \max(\hat{\mathbf{x}}_{i,k}) > p_{min}.
\end{equation}

\subsection{Estimating Excavation Rate}
The previous sections describe the process of updating the global map.
Changes in the global map over time can be used to plan future excavation sequences, make informed decisions about where to dig next, and also allow for real-time productivity monitoring.
An initialization period of $m$ scans is used to allow the global map, $M_{\mathcal{M}}$, to construct a confident belief of the environment.
As mentioned previously, a terrain survey can also be used.

Estimating excavated volume sequentially requires first identifying common cells within the instances being compared.
This is important as the newer instance may have updated a previously unobserved area and does not allow for a fair comparison with the previous instance.
The cell change in volume between times $k_{1}$, and $k_{2}$, is given by,
\begin{equation}
    \label{eq:volume_change}
    c_{i,(k_{1},k_{2})}^{v} = (h_{k_{1}}-h_{k_{2}})\Delta^2 \hspace{1mm} | \hspace{1mm} c_{i} \in M_{\mathcal{M},k_{1}} \cap M_{\mathcal{M},k_{2}}
\end{equation}
The change in volume for an area is calculated by summing the changes in the commonly observed cells.

\subsection{Summary and Limitations}
\begin{figure}[t!]
    \begin{center}
        \includegraphics[width=\linewidth]{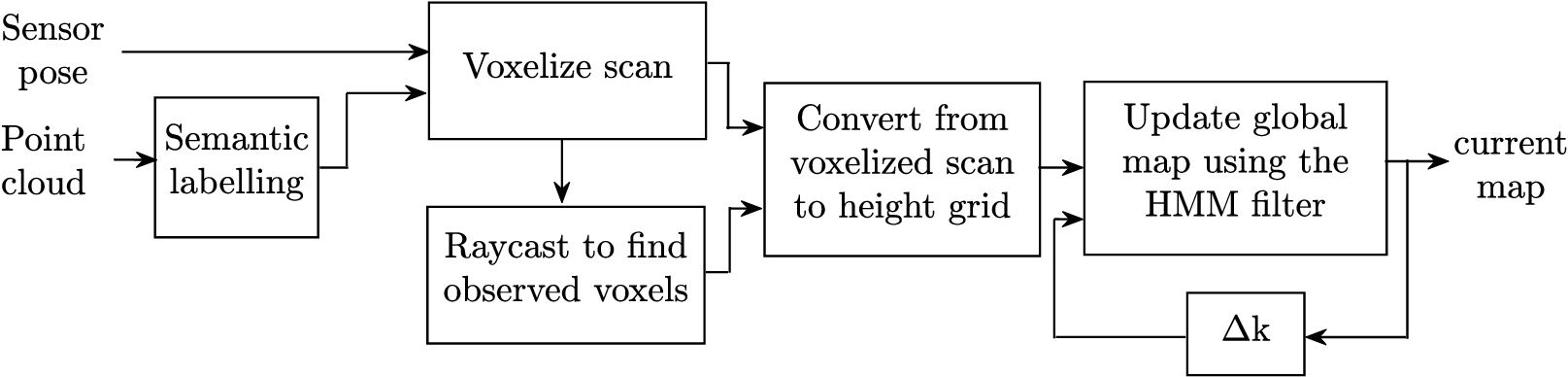}
        \caption{A summary of the map update process. The point cloud is first semantically labelled to estimate points associated with the terrain and then voxelized to find the observed space. This information is used to convert to a height grid and compute the likelihood of each cell being in the $n$ states. The observation is fused with the existing map using the HMM filter.} 
    \label{fig:methodology_flowchart}
    \end{center}
\end{figure}
Figure.~\ref{fig:methodology_flowchart} summarizes the integration of a new observation into the global map.
Modelling the cells as an HMM allows for fast updates of dynamic elements.
The raycasting process helps suppress non-terrain points from being updated to allow for an accurate calculation of material extraction rates.
Limitations of the proposed approach include its dependence on accurate semantic labelling of the point cloud, low-drift sensor odometry, and adequate sensor observability of the terrain.
The height grid representation limits the modelling approach to terrain with no overhang.

\section{Results}
\label{sec:results}
\begin{figure*}[t!]
    \begin{center}
    \begin{subfigure}[b]{0.47\linewidth}
        \includegraphics[width=\linewidth]{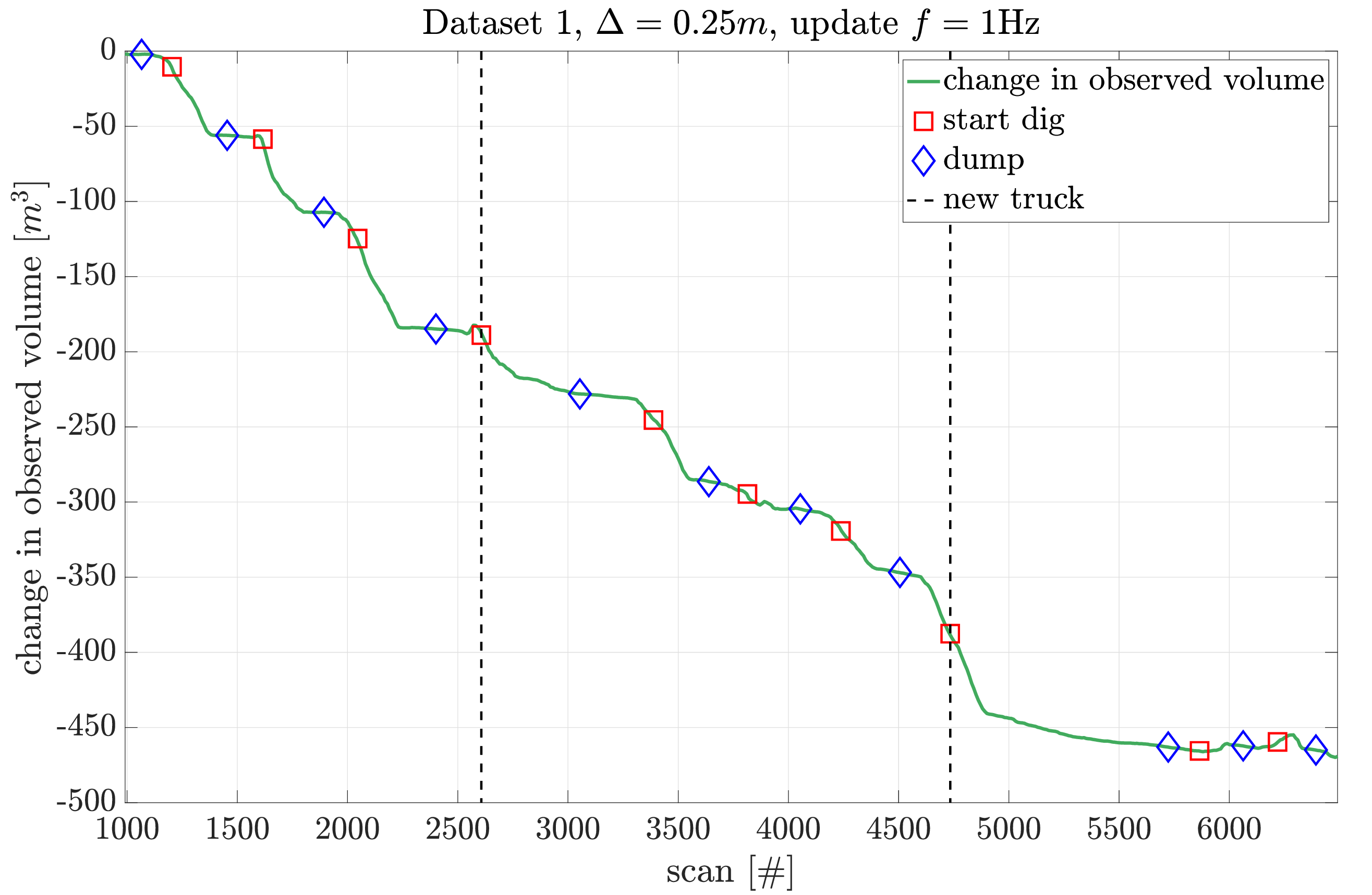} 
    \end{subfigure}
    \begin{subfigure}[b]{0.47\linewidth}
        \includegraphics[width=\linewidth] {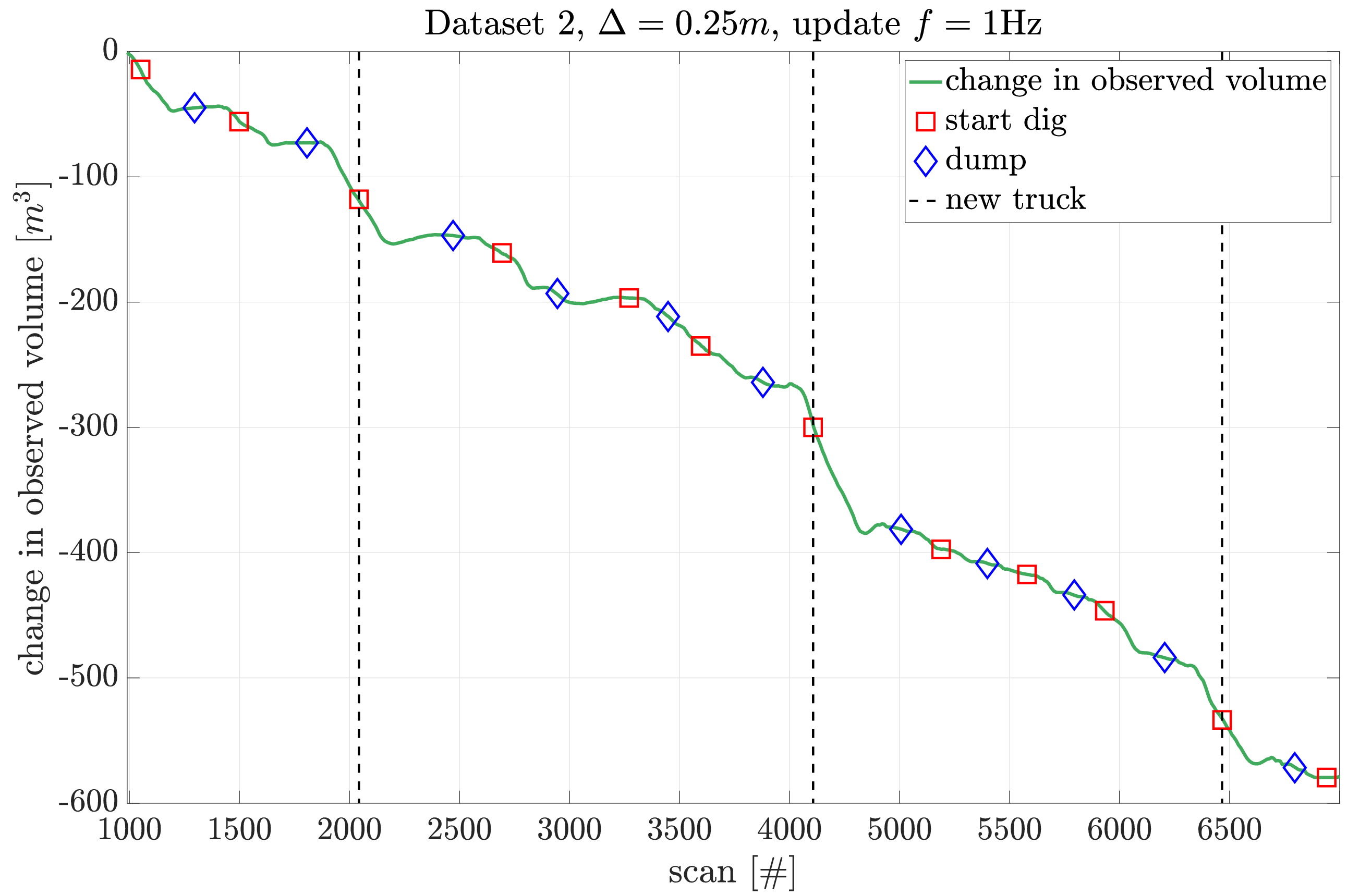}
    \end{subfigure}
    \caption{The net change in observed volume calculated for two datasets.
             The datasets are manually labelled to provide a reference of the shovel's behaviour during the dataset collection.
             Red squares indicate the beginning of a dig and blue diamonds are the dumping of material in a haul truck.}
    \label{fig:volume_results}
    \end{center}
\end{figure*}
The proposed mapping approach has been tested on several datasets recorded using a 3D LiDAR mounted to a CAT 7495 electric rope shovel.
The datasets are recorded during routine operations as the shovel excavates material from the terrain and dumps it into haul trucks.
Results for two datasets consisting of approximately 7000 scans recorded at 10\,Hz are used here.

The algorithm is configured to a map height range of $[h_{min},h_{max}]=[0,20]$\,m, at a discretization size of $\Delta=0.25$\,m, resulting in each cell having $81$ states to represent all possible heights within the desired range.
The standard deviation, $\sigma$, used to construct the likelihood matrix, $\mathbf{B}_{i}$, is set to equal $\Delta$.
The cell size is also set to 0.25m.
The map is initialized as an empty height grid due to the absence of a workspace survey.
Extraction rates are estimated after an initialization period of $m=1000$ scans ($<$ 2\,minutes).

\begin{figure*}[t!]
    \begin{center}
        \includegraphics[width=0.95\linewidth]{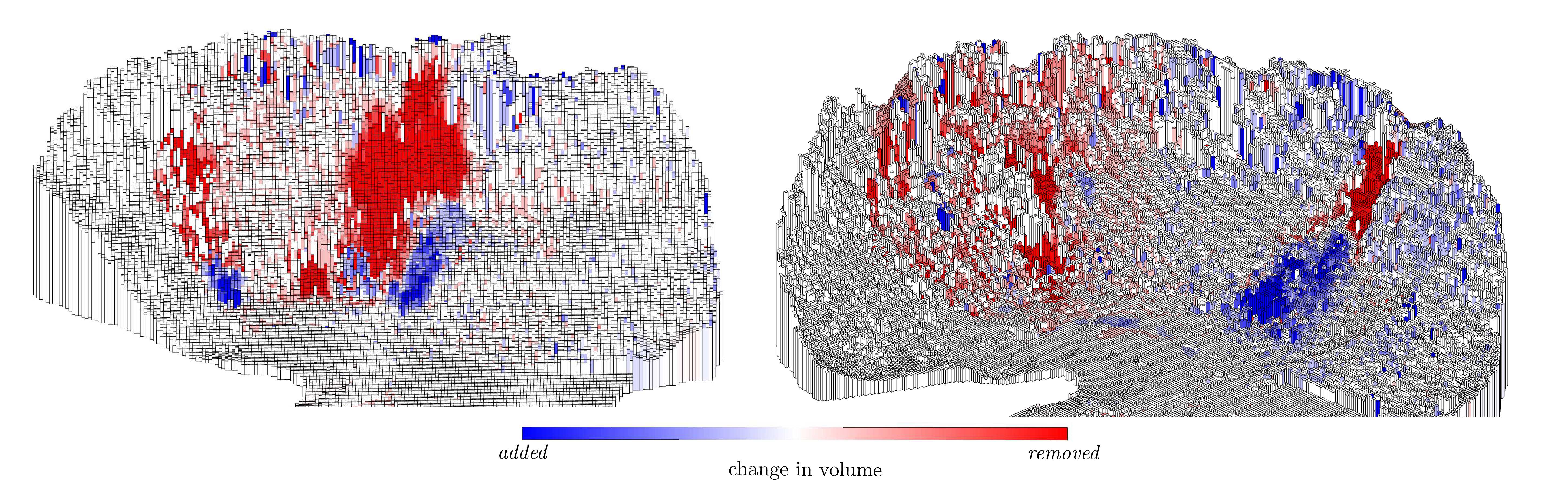}
        \caption{Two examples visualizing the change in the state of the terrain over a receding window of 1000 scans.
        Red cells indicate the removal of material $(h_{new,i} < h_{old,i})$ and blue cells represent the addition of material $(h_{new,i} > h_{old,i})$.
        Both instances show areas of excavation and material spillage.} 
    \label{fig:results_change_in_volume}
    \end{center}
\end{figure*}
Figure.~\ref{fig:shovel_scene} displays a typical scene with a model of the shovel and truck.
The height grid is generated by recursively fusing observations into a global map.
Figure.~\ref{fig:volume_results} displays the extraction rates for the two datasets.
The plots indicate the net change in observed volume in the workspace calculated using (\ref{eq:volume_change}).
We find using 10\,Hz data provides a minimal change in the maps produced, and hence use a lower update frequency of 1\,Hz to reduce CPU load.
The proposed algorithm is capable of providing map updates at 10\,Hz using multithreading on a 12th Gen Intel Core i5-1240P×16 with 16GB RAM.

It is important to note that the changes in volume represented in Fig.~\ref{fig:volume_results} lag the instant the material is excavated due to the limited observability of the workspace and occlusions caused by the shovel as pictured by the point cloud in Fig.~\ref{fig:shovel_scene}.
For example, the shovel swings immediately after filling the dipper, and the sensor does not capture the changed terrain immediately.
The observations are updated when the shovel returns to dig in the same area, leading to a slight lag in updating beliefs.
Sensors with a larger field of view or multiple sensors can help mitigate this issue.
The datasets are manually annotated in Fig.~\ref{fig:volume_results} to provide a reference for the shovel's actions and examine how the volume in the workspace is expected to change.
A ground truth is not available to quantitatively evaluate the mapping accuracy.
The red squares indicate the shovel beginning to dig, and the blue diamonds indicate the dumping of material.
The dashed lines indicate the arrival of a new truck for loading.
Both datasets record a net change of approximately $200\,m^3$ per truck, with each truck being loaded in four cycles.

Example changes in terrain over a receding window of 1000 scans are illustrated in Fig.~\ref{fig:results_change_in_volume}.
Areas with material removed are coloured red, and areas with an addition of material are coloured blue.
Material addition refers to material falling from the digface when disturbed by the dipper, and spillage.
Figure.~\ref{fig:results_change_in_volume} (left) shows two main areas excavated by the shovel in the receding window, with material addition on either side of the area likely caused by spillage from the dipper.
The plot on the right shows material disturbed from the top of the digface and accumulated at the toe.
Accounting for the addition and removal of material allows for a net volume change to be calculated.

\section{Conclusions and Future Work}
\label{sec:conclusions_and_future_work}
This paper introduces a novel approach to probabilistically fuse new observations into an existing map.
Representing each cell using an HMM allows for robustness against spurious measurements from dust and the semantic labelling process.
The results demonstrate real-time terrain mapping using field datasets from a CAT 7495 shovel.

Future work involves recording a survey before and after the test sequence to evaluate the mapping accuracy and provide a quantitative comparison with existing mapping frameworks.
Material movement models can also be used to allow for map updates in areas unobserved by the LiDAR or provide a secondary measurement source to fuse with point cloud data as demonstrated by~\cite{DAd:23}.
This is important when applying the same mapping framework to other mining agents such as dozers, as it may be difficult to capture terrain transformed by the blade in the point cloud data directly.

\small
\bibliography{ifacconf}

\begin{thebibliography}{26}
\providecommand{\natexlab}[1]{#1}
\providecommand{\url}[1]{\texttt{#1}}
\providecommand{\urlprefix}{URL }
\expandafter\ifx\csname urlstyle\endcsname\relax
  \providecommand{\doi}[1]{doi:\discretionary{}{}{}#1}\else
  \providecommand{\doi}{doi:\discretionary{}{}{}\begingroup
  \urlstyle{rm}\Url}\fi

\bibitem[{Bettens(2023)}]{Bet:23}
Bettens, S. (2023).
\newblock \emph{Mission planning under bounded rationality for autonomous
  excavation}.
\newblock Ph.D. thesis, School of Mechanical \& Mining Engineering, The
  University of Queensland.

\bibitem[{Bhandari et~al.(2023)Bhandari, Phillips, and McAree}]{Bha:23}
Bhandari, V., Phillips, T.G., and McAree, P.R. (2023).
\newblock Real-time 6-dof pose estimation of known geometries in point cloud
  data.
\newblock \emph{Sensors}, 23(6).
\newblock \doi{10.3390/s23063085}.

\bibitem[{Bhandari et~al.(2024)Bhandari, Phillips, and McAree}]{Bha:24}
Bhandari, V., Phillips, T.G., and McAree, P.R. (2024).
\newblock Minimal configuration point cloud odometry and mapping.
\newblock \emph{The International Journal of Robotics Research}, 0(0),
  02783649241235325.
\newblock \doi{10.1177/02783649241235325}.

\bibitem[{Dunbabin and Corke(2006)}]{Dun:06}
Dunbabin, M. and Corke, P. (2006).
\newblock Autonomous excavation using a rope shovel.
\newblock \emph{Journal of Field Robotics}, 23(6-7), 379--394.
\newblock \doi{10.1002/rob.20132}.

\bibitem[{D’Adamo(2023)}]{DAd:23}
D’Adamo, T.A. (2023).
\newblock \emph{Building and maintaining real-time terrain maps for autonomous
  bulldozer mining}.
\newblock Ph.D. thesis, School of Mechanical \& Mining Engineering, The
  University of Queensland.

\bibitem[{D’Adamo et~al.(2018)D’Adamo, Phillips, and McAree}]{Dad:18}
D’Adamo, T.A., Phillips, T.G., and McAree, P.R. (2018).
\newblock Registration of three-dimensional scanning lidar sensors: An
  evaluation of model-based and model-free methods.
\newblock \emph{Journal of Field Robotics}, 35(7), 1182--1200.
\newblock \doi{10.1002/rob.21811}.

\bibitem[{Elliott et~al.(2008)Elliott, Aggoun, and Moore}]{Ell:08}
Elliott, R.J., Aggoun, L., and Moore, J.B. (2008).
\newblock \emph{Hidden Markov models: estimation and control}, volume~29.
\newblock Springer Science \& Business Media.
\newblock \doi{10.1007/978-0-387-84854-9}.

\bibitem[{Fan et~al.(2022)Fan, Shen, Chen, Zhang, and Pan}]{Fan:22}
Fan, T., Shen, B., Chen, H., Zhang, W., and Pan, J. (2022).
\newblock Dynamicfilter: an online dynamic objects removal framework for highly
  dynamic environments.
\newblock In \emph{2022 International Conference on Robotics and Automation
  (ICRA)}, 7988--7994.
\newblock \doi{10.1109/ICRA46639.2022.9812356}.

\bibitem[{Hillier et~al.(2015)Hillier, Ryde, and Widzyk-Capehart}]{Hil:15}
Hillier, N., Ryde, J., and Widzyk-Capehart, E. (2015).
\newblock \emph{Comparison of Scanning Laser Range-Finders and Millimeter-Wave
  Radar for Creating a Digital Terrain Map}, 23--38.
\newblock Springer Berlin Heidelberg.
\newblock \doi{10.1007/978-3-662-45514-2\_3}.

\bibitem[{Hornung et~al.(2013)Hornung, Wurm, Bennewitz, Stachniss, and
  Burgard}]{Hor:2013}
Hornung, A., Wurm, K.M., Bennewitz, M., Stachniss, C., and Burgard, W. (2013).
\newblock Octomap: An efficient probabilistic 3d mapping framework based on
  octrees.
\newblock \emph{Autonomous robots}, 34, 189--206.
\newblock \doi{10.1007/s10514-012-9321-0}.

\bibitem[{James et~al.(2024)James, Ford, and Molloy}]{Jam:2024}
James, J., Ford, J.J., and Molloy, T.L. (2024).
\newblock A framework for bayesian quickest change detection in general
  dependent stochastic processes.
\newblock \emph{IEEE Control Systems Letters}, 8, 790--795.
\newblock \doi{10.1109/LCSYS.2024.3403918}.

\bibitem[{Kim and Kim(2020)}]{Kim:20}
Kim, G. and Kim, A. (2020).
\newblock Remove, then revert: Static point cloud map construction using
  multiresolution range images.
\newblock In \emph{2020 IEEE/RSJ International Conference on Intelligent Robots
  and Systems (IROS)}, 10758--10765.
\newblock \doi{10.1109/IROS45743.2020.9340856}.

\bibitem[{Lim et~al.(2021)Lim, Hwang, and Myung}]{Lim:21}
Lim, H., Hwang, S., and Myung, H. (2021).
\newblock Erasor: Egocentric ratio of pseudo occupancy-based dynamic object
  removal for static 3d point cloud map building.
\newblock \emph{IEEE Robotics and Automation Letters}, 6(2), 2272--2279.
\newblock \doi{10.1109/LRA.2021.3061363}.

\bibitem[{Liu et~al.(2019)Liu, Fan, Liu, and Zhang}]{Liu:19}
Liu, K., Fan, Z., Liu, M., and Zhang, S. (2019).
\newblock Object-aware semantic mapping of indoor scenes using octomap.
\newblock In \emph{2019 Chinese Control Conference (CCC)}, 8671--8676.
\newblock \doi{10.23919/ChiCC.2019.8865848}.

\bibitem[{Moravec and Elfes(1985)}]{Mor:85}
Moravec, H. and Elfes, A. (1985).
\newblock High resolution maps from wide angle sonar.
\newblock In \emph{Proceedings. 1985 IEEE International Conference on Robotics
  and Automation}, volume~2, 116--121.
\newblock \doi{10.1109/ROBOT.1985.1087316}.

\bibitem[{Nouiraa et~al.(2016)Nouiraa, Deschaud, and Goulettea}]{Nou:16}
Nouiraa, H., Deschaud, J.E., and Goulettea, F. (2016).
\newblock Point cloud refinement with a target-free intrinsic calibration of a
  mobile multi-beam lidar system.
\newblock \emph{The International Archives of the Photogrammetry, Remote
  Sensing and Spatial Information Sciences}, XLI-B3, 359--366.
\newblock \doi{10.5194/isprs-archives-XLI-B3-359-2016}.

\bibitem[{Phillips et~al.(2016)Phillips, Green, and McAree}]{Phi:16}
Phillips, T.G., Green, M.E., and McAree, P.R. (2016).
\newblock Is it what i think it is? is it where i think it is? using
  point-clouds for diagnostic testing of a digging assembly's form and pose for
  an autonomous mining shovel.
\newblock \emph{Journal of Field Robotics}, 33(7), 1013--1033.
\newblock \doi{10.1002/rob.21643}.

\bibitem[{Phillips et~al.(2017)Phillips, Guenther, and McAree}]{Phi:2017}
Phillips, T.G., Guenther, N., and McAree, P.R. (2017).
\newblock When the dust settles: The four behaviors of lidar in the presence of
  fine airborne particulates.
\newblock \emph{Journal of field robotics}, 34(5), 985--1009.
\newblock \doi{10.1002/rob.21701}.

\bibitem[{Phillips and McAree(2018)}]{Phi:18}
Phillips, T.G. and McAree, P.R. (2018).
\newblock An evidence-based approach to object pose estimation from lidar
  measurements in challenging environments.
\newblock \emph{Journal of Field Robotics}, 35(6), 921--936.
\newblock \doi{10.1002/rob.21788}.

\bibitem[{Rabiner(1989)}]{Rab:89}
Rabiner, L. (1989).
\newblock A tutorial on hidden markov models and selected applications in
  speech recognition.
\newblock \emph{Proceedings of the IEEE}, 77(2), 257--286.
\newblock \doi{10.1109/5.18626}.

\bibitem[{Souza et~al.(2013)Souza, Maia, Aroca, and Gonçalves}]{Sou:13}
Souza, A., Maia, R.S., Aroca, R.V., and Gonçalves, L.M.G. (2013).
\newblock Probabilistic robotic grid mapping based on occupancy and elevation
  information.
\newblock In \emph{2013 16th International Conference on Advanced Robotics
  (ICAR)}, 1--6.
\newblock \doi{10.1109/ICAR.2013.6766467}.

\bibitem[{Thrun et~al.(1998)Thrun, Burgard, and Fox}]{Thr:98}
Thrun, S., Burgard, W., and Fox, D. (1998).
\newblock A probabilistic approach to concurrent mapping and localization for
  mobile robots.
\newblock \emph{Autonomous Robots}, 5, 253--271.
\newblock \doi{10.1023/A:1007436523611}.

\bibitem[{Thrun et~al.(2000)Thrun, Burgard, and Fox}]{Thr:00}
Thrun, S., Burgard, W., and Fox, D. (2000).
\newblock A real-time algorithm for mobile robot mapping with applications to
  multi-robot and 3d mapping.
\newblock In \emph{Proceedings 2000 ICRA. Millennium Conference. IEEE
  International Conference on Robotics and Automation. Symposia Proceedings
  (Cat. No.00CH37065)}, volume~1, 321--328 vol.1.
\newblock \doi{10.1109/ROBOT.2000.844077}.

\bibitem[{Yguel et~al.(2008)Yguel, Aycard, and Laugier}]{Ygu:08}
Yguel, M., Aycard, O., and Laugier, C. (2008).
\newblock \emph{Update Policy of Dense Maps: Efficient Algorithms and Sparse
  Representation}, 23--33.
\newblock Springer Berlin Heidelberg, Berlin, Heidelberg.
\newblock \doi{10.1007/978-3-540-75404-6\_3}.

\bibitem[{Yin et~al.(2024)Yin, Xu, Lu, Chen, Xiong, Shen, Stachniss, and
  Wang}]{Yin:24}
Yin, H., Xu, X., Lu, S., Chen, X., Xiong, R., Shen, S., Stachniss, C., and
  Wang, Y. (2024).
\newblock A survey on global lidar localization: Challenges, advances and open
  problems.
\newblock \emph{International Journal of Computer Vision}, 1--33.
\newblock \doi{10.1007/s11263-024-02019-5}.

\bibitem[{Zhang et~al.(2018)Zhang, Wei, Shen, Wei, Zhu, and Song}]{Zha:18}
Zhang, L., Wei, L., Shen, P., Wei, W., Zhu, G., and Song, J. (2018).
\newblock Semantic slam based on object detection and improved octomap.
\newblock \emph{IEEE Access}, 6, 75545--75559.
\newblock \doi{10.1109/ACCESS.2018.2873617}.

\end{thebibliography}

\end{document}